\documentclass[11pt,a4paper]{article}
\usepackage[hyperref]{emnlp-ijcnlp-2019}
\usepackage{times}
\usepackage{latexsym}

\usepackage{url}
\usepackage{amsmath}
\usepackage{graphicx}
\aclfinalcopy 

\title{WSLLN: Weakly Supervised Natural Language Localization Networks}

\author{
    Mingfei Gao$^1$~\thanks{Work done
    when the author was at Salesforce Research.}
    \hspace{0.35cm} Larry S. Davis$^1$
    \hspace{0.25cm} Richard Socher$^2$
    \hspace{0.25cm} Caiming Xiong$^2$~\thanks{Corresponding author.}\\[.6ex]
    $^1$University of Maryland
    \hspace{0.25cm} $^2$Salesforce Research \\
    {\tt\small  \{mgao,lsd\}@umiacs.umd.edu, \{rsocher,cxiong\}@salesforce.com}
}
\date{}

\begin{document}
\maketitle
\begin{abstract}
We propose weakly supervised language localization networks (WSLLN) to detect events in long, untrimmed videos given language queries. To learn the correspondence between visual segments and texts, most previous methods require temporal coordinates (start and end times) of events for training, which leads to high costs of annotation. WSLLN relieves the annotation burden by training with only video-sentence pairs without accessing to temporal locations of events. With a simple end-to-end structure, WSLLN measures segment-text consistency and conducts segment selection (conditioned on the text) simultaneously. Results from both are merged and optimized as a video-sentence matching problem. Experiments on \emph{ActivityNet Captions} and \emph{DiDeMo} demonstrate that WSLLN achieves state-of-the-art performance.
\end{abstract}

\section{Introduction}
Extensive work has been done on temporal action/activity localization~\cite{shou2016temporal,zhao2017temporal,dai2017temporal,buch2017sst,gao2017turn,chao2018rethinking}, where an action of interest is segmented from long, untrimmed videos. These methods only identify actions from a pre-defined set of categories, which limits their application to situations where only unconstrained language descriptions are available. This more general problem is referred to as natural language localization (NLL)~\cite{anne2017localizing,gao2017tall}. The goal is to retrieve a temporal segment from an untrimmed video based on an arbitrary text query. Recent work focuses on learning the mapping from visual segments to the input text~\cite{anne2017localizing,gao2017tall,liu2018temporal,hendricks18emnlp,zhang2018man} and retrieving segments based on the alignment scores. However, in order to successfully train a NLL model, a large number of diverse language descriptions are needed to describe different temporal segments of videos which incurs high human labeling cost.

We propose Weakly Supervised Language Localization Networks (WSLLN) which requires only video-sentence pairs during training with no information of where the activities temporally occur. Intuitively, it is much easier to annotate video-level descriptions than segment-level descriptions. Moreover, when combined with text-based video retrieval techniques, video-sentence pairs may be obtained with minimum human intervention. The proposed model is simple and clean, and can be trained end-to-end in a single stage. We validate our model on \emph{ActivityNet Captions} and \emph{DiDeMo}. The results show that our model achieves the state-of-the-art of the weakly supervised approach and has comparable performance as some supervised approaches.

\section{Related Work}
\noindent\textbf{Temporal Action Localization} in long videos is widely studied in both offline and online scenarios. In the offline setting, temporal action detectors~\cite{shou2016temporal,buch2017sst,gao2017turn,chao2018rethinking} predict the start and end times of actions after observing the whole video, while online approaches~\cite{de2016online,gao2017red,shou2018online,xu2018temporal,gao2019startnet} label action class in a per-frame manner without accessing future information. The goal of temporal action detectors is to localize actions in pre-defined categories. However, activities in the wild is very complicated and it is challenging to cover all the activities of interest by using a finite set of categories.

\noindent\textbf{Natural Language Localization} in untrimmed videos was first introduced in~\cite{gao2017tall,anne2017localizing}, where given an arbitrary text query, the methods attempt to localize the text (predict its start and end times) in a video. Hendricks \emph{et al.} proposed MCN~\cite{anne2017localizing} which embeds the features of visual proposals and sentence representations in the same space and ranks proposals according their similarity with the sentence. Gao \emph{et al.} proposed CTRL~\cite{gao2017tall}, where alignment and regression are conducted for clip candidates. Liu \emph{et al.} introduced TMN~\cite{liu2018temporal} which measures the clip-sentence alignment guided by the semantic structure of the text query. Later, Hendricks \emph{et al.} proposed MLLC~\cite{hendricks18emnlp} that explicitly reasons about temporal clips of a video. Zhang \emph{et al.} proposed MAN~\cite{zhang2018man} which utilizes Graph Convolutional Networks~\cite{kipf2016semi} to model temporal relations among visual clips. Although these methods achieve considerable success, they need segment-level annotations for training. Duan \emph{et al.} proposed WSDEC to handle weakly supervised dense event captioning in~\cite{duan2018weakly} by alternating between language localization and caption generation iteratively. WSDEC generates language localization as intermediate results and can be trained using video-level labels. Thus, we set it as a baseline, although it is not designed for NLL.

\noindent\textbf{Weakly Supervised Localization} has been studied extensively to use weak supervisions for object detection on images and action localization in videos~\cite{oquab2015object,Bilen_2016_CVPR, tang2017multiple, gao2018c,kantorov2016contextlocnet, Li_2016_CVPR, jie2017deep, diba2017weakly, papadopoulos2017training, duchenne2009automatic,laptev2008learning,bojanowski2014weakly, huang2016connectionist,wang2017untrimmednets,shou2018autoloc}. Some methods use class labels to train object detectors. Oquab \emph{et al.} discussed that object locations may be freely obtained when training classification models~\cite{oquab2015object}. Bilen \emph{et al.} proposed WSDDN~\cite{Bilen_2016_CVPR}, which focuses on both object recognition and localization. Their proposed two-stream architecture inspired several weakly supervised approaches~\cite{tang2017multiple, gao2018c, wang2017untrimmednets} including our method. Li \emph{et al.} presented an adaptation strategy in~\cite{Li_2016_CVPR} which uses the output of a weak detector as pseudo groundtruth to train a detector in a fully supervised way. OICR~\cite{tang2017multiple} integrates multiple instance learning and iterative classifer refinement in a single network. Some works use other types of weak supervisions to optimize detectors. In~\cite{papadopoulos2017training}, Papadopoulos \emph{et al.} used clicks to train detectors. Gao \emph{et al.} utilized object counts for weakly supervised object detection~\cite{gao2018c}. Instead of using temporally labeled segments, weakly supervised action detectors use weaker annotations, \emph{e.g.}, movie script~\cite{duchenne2009automatic,laptev2008learning}, the order of the occurring action
classes in videos~\cite{bojanowski2014weakly, huang2016connectionist} and video-level class labels~\cite{wang2017untrimmednets,shou2018autoloc}. 

\section{Weakly Supervised Language Localization Networks (WSLLN)}
\subsection{Problem Statement}
Following the setting of its strongly supervised counterpart~\cite{gao2017tall,anne2017localizing}, the goal of a weakly supervised language localization (WSLL) method is to localize the event that is described by a sentence query in a long, untrimmed video. Formally, given a video consisting of a sequence of image frames, $\textbf{V}_i=[I_i^1, I_i^2, ..., I_i^T]$, and a text query ${Q}_i$, the model aims to localize a temporal segment, $[I_i^{st}, ...,I_i^{ed}]$, which semantically aligns best with the query. $st$ and $ed$ indicate the start and end times, respectively. The difference is that WSLL methods only utilize video-sentence pairs, $\{\textbf{V}_i, Q_i\}_{i=1}^N$, for training, while supervised approaches have access to the start and end times of the queries.
\begin{figure*}[t]
\begin{center}
\includegraphics[width=1.0\linewidth]{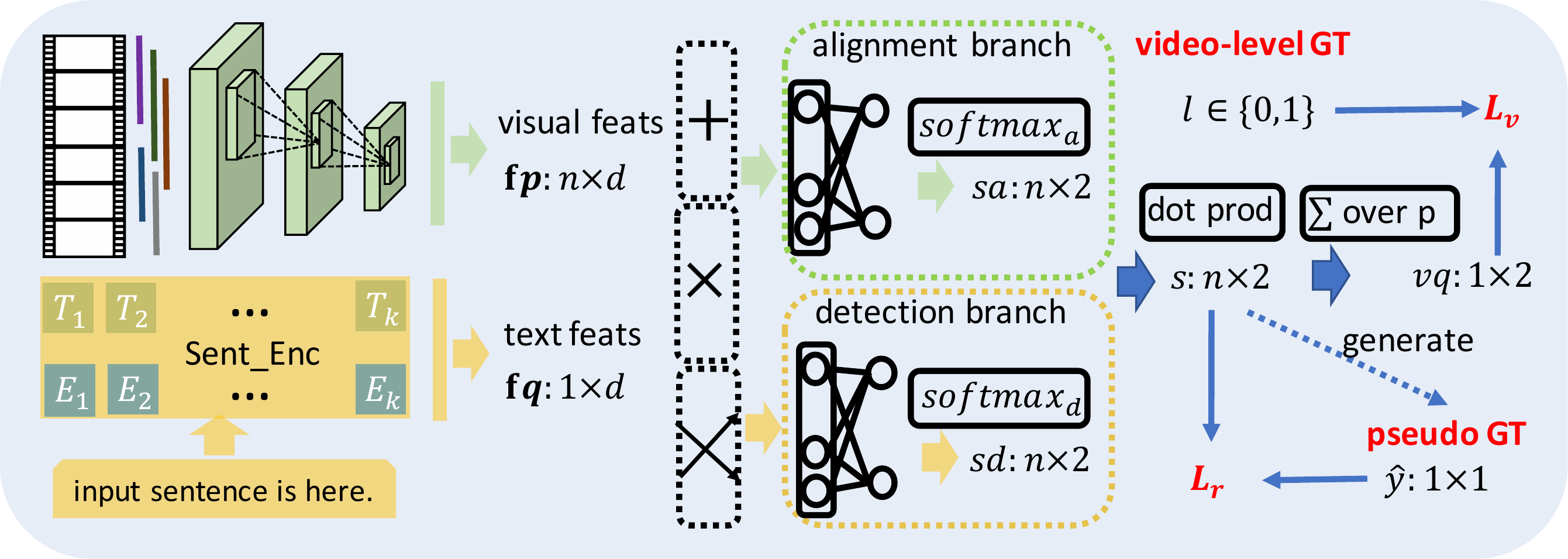}
\end{center}
\caption{
The workflow of our method. Visual and text features are extracted from $n$ video proposals and the input sentence. Fully-connected (FC) layers are used to transform the features to the same length, $d$.  The two features are combined by multi-modal processing~\cite{gao2017tall}  and input to the two-branch structure. Scores from both parts are merged. Video-level scores, $vq$, are obtained by summing $s$ over proposals. The whole pipeline is trained end-to-end using video-level and pseudo segment-level labels. $x\times z$ indicates dimensions.
}
\label{fig:pipeline}
\end{figure*}
\subsection{The Proposed Approach}
Taking frame sequences, $[I_i^1, I_i^2, ..., I_i^T]$, as inputs, the model first generates a set of temporal proposals, $\{p_i^1, p_i^2,...,p_i^n\}$, where $p_i^j$ consists of temporally-continuous image frames. Then, the method aligns the proposals with the input query and outputs scores for proposals, $\{s_i^1, s_i^2,...,s_i^n\}$, indicating their likelihood  of containing the event. 

\noindent\textbf{Feature Description}.
Given a sentence query ${Q}_i$ of arbitrary length, sentence encoders can be used to extract text feature, $fq_i$, from the query. For a video, $\textbf{V}_i=[I_i^1, I_i^2, ..., I_i^T]$, features, $\textbf{fv}_i=[fv_i^1, fv_i^2, ..., fv_i^T]$, are extracted from each frame. Following~\cite{anne2017localizing}, the visual feature, $fp_i^j$, of a proposal $p_i^j$ is obtained using Eq.~\ref{eq: proposal_feat}, where $pool(x, t_1, t_2)$ means average pooling features $x$ from time $t_1$ to $t_2$, $||$ indicates concatenation, $j_{st}$/$j_{ed}$ indicates start/end times of the proposal and $\bar{j}$ means time is normalized to $[0, 1]$.
\begin{equation}
\label{eq: proposal_feat}
pool({\textbf{fv}_i}, j_{st}, j_{ed}) || pool({\textbf{fv}_i}, 0, T)||[\bar{j}_{st}, \bar{j}_{ed}]
\end{equation}

We see that the feature of each proposal contains the information of its visual pattern, the overall context and its relative position in the video.

Following~\cite{gao2017tall}, features of the sentence and a visual proposal are combined as in Eq.~\ref{eq: vs_comb}. The feature, $fm$, will be used to measure the matching between a candidate proposal and the input query.
\begin{equation}
\label{eq: vs_comb}
fm=(fp+fq) || (fp\cdot fq) || FC(fp||fq)
\end{equation}

The workflow of WSLLN is illustrated in Fig.~\ref{fig:pipeline}. Inspired by the success of the two-stream structure in the weakly supervised object and action detection tasks~\cite{Bilen_2016_CVPR, wang2017untrimmednets}, WSLLN consists of two branches, \emph{i.e.}, alignment branch and selection branch. The semantic consistency between the input text and each visual proposal is measured in the alignment branch. The proposals are compared and selected in the detection branch. Scores from both branches are merged to produce the final results.

\noindent\textbf{Alignment Branch} produces the consistency scores, $sa_i \in \mathrm{R}^{n\times2}=[sa_i^1, sa_i^2,...,sa_i^n]$, for proposals of the video-sentence pair. $sa_i$ in Eq.~\ref{eq: align}, measures how well each proposal matches the text. Different proposal scores are calculated independently where $softmax_a$ indicates applying the softmax function over the last dimension. 

\begin{equation}
\label{eq: align}
sa_i=softmax_a(\textbf{W}_afm_i)
\end{equation}
\noindent\textbf{Detection Branch} performs proposal selection. The selection score, $sd_i \in \mathrm{R}^{n\times2}=[sd_i^1, sd_i^2,...,sd_i^n]$ in Eq.~\ref{eq: detection}, is obtained by applying softmax function over proposals. Through softmax, the score of a proposal will be affected by those of other
proposals, so this operation encourages competition among segments. 

\begin{equation}
\label{eq: detection}
sd_i=softmax_d(\textbf{W}_dfm_i)
\end{equation}
\noindent\textbf{Score Merging} is applied to both parts to obtain the results by dot production, \emph{i.e.}, $s_i=sa_i \cdot sd_i$, for proposals. $s_i$ is used as the final segment-sentence matching scores during inference.

\noindent\textbf{Training Phase}. To utilize video-sentence pairs as supervision, our model is optimized as a video-sentence matching classifier. We compute the matching score of a given video-sentence pair by summing $s_i^j$ over proposals, $vq_i=\sum_{j=1}^{n} s_i^j$. Then, $L_v$ is obtained in Eq.~\ref{eq: lv} by measuring the score with the video-sentence match label $l_i \in \{0, 1\}$. Positive video-sentence pairs can be obtained directly. We generate negative ones by pairing each video with a randomly selected sentence in the training set. We ensure that the positive pairs are not included in the negative set.

\begin{equation}
\label{eq: lv}
L_v=loss(vq_i, l_i)
\end{equation}

Results can be further refined by adding an auxiliary task $ L_r$ in Eq.~\ref{eq: lr} where $\hat{y}_i=\{0, 1, ..., n-1 \}$ indicates the index of the segment that best matches the sentence during training. The real segment-level labels are not available, thus we generate pseudo labels by setting $\hat{y}_i={\operatorname*{argmax}}_{j} s_i^j[:,1]$. This loss further encourages competition among proposals.

\begin{equation}
\label{eq: lr}
L_r=loss(s_i^j, \hat{y}_i)
\end{equation}

The overall objective is minimizing $L$ in Eq.~\ref{eq: loss_full}, where $\lambda$ is a balancing scalar. $loss$ is cross-entropy loss.

 \begin{equation}
 \label{eq: loss_full}
    L= loss(vq_i, l_i) +\lambda loss(s_i^j, \hat{y}_i) .
\end{equation}

\section{Experiments}
\subsection{Experimental Settings}
\noindent\textbf{Implementation Details}. BERT~\cite{devlin2018bert} is used as the sentence encoder, where the feature of `[CLS]' at the last layer is extracted as the sentence representation. Visual and sentence features are linearly transformed to have the same dimension, $d=1000$. The hidden layers for both branches have 256 units. For \emph{ActivityNet Captions}, we take the $n=15$ proposals over multiple scales of each video provided by~\cite{duan2018weakly} and use the C3D~\cite{tran2015learning} features provided by~\cite{krishna2017dense}. For \emph{DiDeMo}, we use the $n=21$ proposals and VGG~\cite{simonyan2014very} features (RGB and Flow) provided in~\cite{anne2017localizing}.

\noindent\textbf{Evaluation Metrics}. Following~\cite{gao2017tall,anne2017localizing}, \emph{R@k,IoU=th} and \emph{mIoU} are used for evaluation. Proposals are ranked according to their matching scores with the input sentence. If the temporal IoU between at least one of the top-k proposals and the groundtruth is bigger or equal to $th$, the sentence is counted as matched. \emph{R@k,IoU=th} means the percentage of matched sentences over the total sentences given $k$ and $th$. \emph{mIoU} is the mean IoU between the top-1 proposal and the groundtruth. 
\subsection{Experiments on ActivityNet Captions}
\noindent\textbf{Dataset Description}. \emph{ActivityNet Captions}~\cite{krishna2017dense} is a large-scale dataset of human activities. It contains 20k videos including 100k video-sentences in total. We train our models on the training set and test them on the validation set. Although the dataset provides segment-level annotation, we only use video-sentence pairs during training.

\noindent\textbf{Baselines}. We compare with strongly supervised approaches, \emph{i.e.}, CTRL~\cite{gao2017tall}, ABLR~\cite{yuan2018find} and WSDEC-S~\cite{duan2018weakly} to see how much accuracy it sacrifices when using only weak labels. Originally proposed for dense-captioning, WSDEC-W~\cite{duan2018weakly} achieves state-of-the-art performance for weakly supervised language localization. Although showing good performance, WSDEC-W involves complicated training stages, and alternates between sentence localization and caption generation for iterations.
\begin{table}[]
\centering
\setlength\tabcolsep{1.7pt}
\begin{tabular}{lccccc}
Model & WS & IoU=0.1 & IoU=0.3 & IoU=0.5 & mIoU \\
\hline
CTRL & False & 49.1 & 28.7 & 14.0 & 20.5 \\
ABLR& False & 73.3 & 55.7 & 36.8 & 37.0\\
WSDEC-S & False & 70.0 & 52.9 & 37.6 & 40.4 \\
\hline
WSDEC-W & True & 62.7 & 42.0 & 23.3 & 28.2 \\
\textbf{WSLLN} & True &75.4  &42.8  &22.7  & 32.2\\
\hline
\end{tabular}
\caption{Comparison results based on $R@1$ on \emph{ActivityNet Captions}. All baseline numbers are reprinted from~\cite{duan2018weakly}. WS: weakly supervised.}
\label{tab: act}
\end{table}

\begin{table}[]
\centering
\begin{tabular}{lcccccc}
 $\lambda$ $\rightarrow$& 0.0 & 0.1 & 0.2 & 0.3 & 0.4 & 0.5 \\
 \hline
IoU=0.1 & 64.9 &75.4&  75.5&  75.5&  75.5&  66.6  \\
IoU=0.3 &  36.2&42.8&  42.9&  42.9&  42.9&  38.3  \\
IoU=0.5 &  19.4&22.7&  22.7&  22.8&  22.7&  20.7 \\
mIoU &  27.4& 32.2 &  32.3&  32.3&  32.3& 28.8 \\
\hline
\end{tabular}
\caption{R@1 results of our method on \emph{ActivityNet Captions} when $\lambda$ in Eq.~\ref{eq: loss_full} is set to be different values.}
\label{tab: ablation_act}
\end{table}
\subsubsection{ Comparison Results} 
Comparison results are displayed in Tab.~\ref{tab: act}. It shows that WSLLN largely outperforms WSDEC-W by $\sim$$4\%$ $mIoU$. When comparing with strongly supervised methods, WSLLN outperforms CTRL by over $11\%$ $mIoU$. Using the $R@1, IoU=0.1$ metric, our model largely outperforms all the baselines including strongly and weakly supervised methods which means that when a scenario is flexible with the IoU coverage, our method has great advantage over others. When $th=$$0.3/0.5$, our model has comparable results as WSDEC-W and largely outperforms CTRL. The overall results demonstrate good performance of WSLLN, even though there is still a big gap between weakly supervised methods and some supervised ones, \emph{i.e.}, ABLR and WSDEC-S. $mIoU$ (mean$\pm$std) of WSLLN across 3 runs is $32.2\pm0.05$ which demonstrates the robustness of our method.

\subsubsection{Ablation Study} 
\noindent\textbf{Effect of $\lambda$}. We evaluate the effect of $\lambda$ (see  Eq.~\ref{eq: loss_full}) in Tab.~\ref{tab: ablation_act}. As it shows, our model performs stable when $\lambda$ is set from $0.1$ to $0.4$. When $\lambda=0$, the refining module is disabled and the performance drops. When $\lambda$ is set to a big number, \emph{e.g.}, $0.5$, the contribution of $L_v$ is reduced and the model performance also drops.

\noindent\textbf{Effect of Sentence Encoder}. WSDEC-W uses GRU~\cite{cho2014learning} as its sentence encoder, while our method uses BERT. It seems an unfair comparison, since BERT is powerful than GRU in general. However, we uses pretrained BERT model without fine tuning on our dataset, while WSDEC-W
uses GRU but performed an end-to-end training. So, it is unclear which setting is better. To resolve this concern, we replace our BERT with GRU following WSDEC-W. The $R@1$ results when $IoU$ is set to be 0.1, 0.3 and 0.5 are 74.0, 42.3 and 22.5, respectively. The mIoU is 31.8. It shows that our model with GRU has comparable results as that with BERT. 

\noindent\textbf{Effect of Two-branch Design}. We create two baselines, \emph{ie}, \emph{Align-only} and \emph{Detect-only}, to demonstrate the effectiveness of our design. To perform fair comparison, both of them are trained using only video-sentence pairs.

\emph{Align-only} contains only the alignment branch. For positive video sentence pair, we give positive labels to all proposals. Negative pairs have negative labels for all the proposals. Loss is calculated between
proposal scores and the generated segment-level labels.

\emph{Detect-only} contains only the detection branch. Loss is calculated using the highest detection score over
proposals and the video-level label at each training iteration.

Comparison results are displayed in Tab.~\ref{tab: act_ablation}. It shows that the two baselines underperform WSLLN by a large margin, which demonstrates
the effectiveness of our design.

\begin{table}[]
\centering
\setlength\tabcolsep{1.7pt}
\begin{tabular}{lcccc}
Model &  IoU=0.1 & IoU=0.3 & IoU=0.5 & mIoU \\
\hline
Align-only & 40.0& 18.9& 7.5& 13.4 \\
Detect-only & 33.7& 18.3& 10.4& 13.6\\
\hline
\end{tabular}
\caption{Ablation study based on $R@1$ on \emph{ActivityNet Captions}. Both methods are trained using weak supervisions.}
\label{tab: act_ablation}
\end{table}
\subsection{Experiments on DiDeMo}
\noindent\textbf{Dataset Description}. \emph{DiDeMo} was proposed in~\cite{anne2017localizing} for the language localization task. It contains 10k, 30-second videos including 40k annotated segment-sentence pairs. Our models are trained using video-sentence pairs in the train set and tested on the test set.

\noindent\textbf{Baselines}. To the best of our knowledge, no weakly supervised method has been evaluated on \emph{DiDeMo}. So, we compare with some supervised methods, \emph{i.e.}, MCN~\cite{anne2017localizing} and LOR~\cite{hu2016natural}. MCN is a supervised NLL model. LOR is a supervised language-object retrieval model. It utilizes much more expensive (object-level) annotations for training. We follow the same setup of LOR as in~\cite{anne2017localizing} to evaluate LOR for our task.

\noindent\textbf{Comparison Results} are shown in Tab.~\ref{tab: didemo}. WSLLN performs better than LOR in terms of $R@1/5$. We also observe that the gap between our method and the supervised NLL model is much larger on \emph{DiDeMo} than on \emph{ActivityNet Captions}. This may be due to the fact that \emph{DiDeMo} is a much smaller dataset which is a disadvantage for weakly supervised learning. 
\begin{table}[]
\centering
\begin{tabular}{lccccc}
Model & WS&Input & R@1 & R@5& mIoU \\
\hline
Chance & --&-- & 3.75 & 22.50 & 22.64 \\
LOR & False &RGB&16.2 &43.9 &27.2 \\
MCN & False& RGB & 23.1 & 73.4 & 35.5 \\
MCN  & False& Flow & 25.8 & 75.4 & 38.9 \\
\hline
\textbf{WSLLN} & True& RGB & 19.4 & 53.1 & 25.4 \\
\textbf{WSLLN} & True & Flow & 18.4 & 54.4 & 27.4 \\
\hline
\end{tabular}
\caption{Comparison results on \emph{DiDeMo}. Following MCN, we set $th=1.0$ for the IoU threshold. All baseline numbers are reprinted from~\cite{anne2017localizing}. WS: weakly supervised.}

\label{tab: didemo}
\end{table}
\section{Conclusion}
We propose WSLLN-- a simple language localization network. Unlike most existing methods which require segment-level supervision, our method is optimized using video-sentence pairs. WSLLN is based on a two-branch architecture where one branch performs segment-sentence alignment and the other one conducts segment selection.  Experiments show that WSLLN achieves promising results on \emph{ActivityNet Captions} and \emph{DiDeMo}.
\bibliography{wsll}
\bibliographystyle{acl_natbib}
\end{document}